\documentclass[10pt]{article}

\usepackage{placeins}
\usepackage{microtype}
\usepackage{times}
\usepackage{graphicx}
\usepackage{subfigure}
\usepackage{algpseudocode}
\usepackage{booktabs} % for professional tables
\usepackage{pgfplots}
\usepackage{bbm}
\usepackage{enumerate}
\usepackage{amsmath}
\usepackage{amsthm}
\usepackage{amssymb}
\usepackage{amsfonts}
\usepackage{mathrsfs}
\usepackage{mathtools}
\usepackage{graphicx}
\usepackage{caption}

\usepackage{nicefrac}

\usepackage[hidelinks]{hyperref}
\usepackage{xcolor}
\usepackage{url}

\newcommand{\pa}[1]{ \left({#1}\right) }
\newcommand{\ha}[1]{ \left[{#1}\right] }
\newcommand{\ca}[1]{ \left\{{#1}\right\} }
\newcommand{\inner}[1]{\left\langle #1 \right\rangle}

\newcommand{\norm}[1]{\left\lVert #1 \right\rVert}
\newcommand{\card}[1]{\left\lvert{#1}\right\rvert}
\newcommand{\abs}[1]{\card{#1}}

\newcommand{\vv}{\textbf{v}}

\newcommand{\vx}{\textbf{x}}

\newcommand{\vzero}{\textbf{0}}

\newcommand{\R}{\mathbb{R}}

\newcommand{\mcA}{\mathcal{A}}

\newcommand{\mcD}{\mathcal{D}}

\newcommand{\mcS}{\mathcal{S}}

\DeclareMathOperator*{\MSE}{MSE}

\newcommand{\indicator}{\mathbbm{1}}

\DeclareMathOperator{\Uniform}{Uniform}

\renewcommand{\d}[1]{\mathop{\mathrm{d} #1 }}
\DeclarePairedDelimiterX{\infdivx}[2]{(}{)}{ #1\;\delimsize\|\;#2 }

\makeatletter
\newcommand{\distas}[1]{\mathbin{\overset{#1}{\kern\z@\sim}}}%
\makeatother

\DeclareMathOperator\mathExp{\mathbb{E}}
\DeclareMathOperator*\mathExpUnder{\mathbb{E}}
\newcommand{\E}{\mathExp}

\DeclareMathOperator*{\argmin}{argmin}
\DeclareMathOperator*{\argmax}{argmax}

\renewcommand{\d}[1]{\mathop{\mathrm{d} #1 }}

\makeatletter
\DeclareFontFamily{U}  {MnSymbolF}{}
\DeclareSymbolFont{symbolsMN}{U}{MnSymbolF}{m}{n}
\SetSymbolFont{symbolsMN}{bold}{U}{MnSymbolF}{b}{n}
\DeclareFontShape{U}{MnSymbolF}{m}{n}{
    <-6>  MnSymbolF5
   <6-7>  MnSymbolF6
   <7-8>  MnSymbolF7
   <8-9>  MnSymbolF8
   <9-10> MnSymbolF9
  <10-12> MnSymbolF10
  <12->   MnSymbolF12}{}
\DeclareFontShape{U}{MnSymbolF}{b}{n}{
    <-6>  MnSymbolF-Bold5
   <6-7>  MnSymbolF-Bold6
   <7-8>  MnSymbolF-Bold7
   <8-9>  MnSymbolF-Bold8
   <9-10> MnSymbolF-Bold9
  <10-12> MnSymbolF-Bold10
  <12->   MnSymbolF-Bold12}{}
\DeclareMathSymbol{\tbigtimes}{\mathop}{symbolsMN}{2}
\newcommand*{\bigtimes}{%
  \DOTSB
  \tbigtimes
  \slimits@ 
}
\makeatother

\usepackage[utf8]{inputenc}
\usepackage{tikz}
\usepackage{wrapfig}
\usetikzlibrary{shapes.geometric}

\usepackage{amsthm}
\newtheorem{dfn}{Definition}[section]
\newtheorem{thm}{Theorem}[section]

\newcommand{\figref}[1]{Fig.~\ref{#1}}
\newcommand{\eqnref}[1]{Eq.~(\ref{#1})}

\newcommand{\dfnref}[1]{Definition~\ref{#1}}

\pgfplotsset{every axis/.append style={
    axis x line=middle,    % put the x axis in the middle
    axis y line=middle,    % put the y axis in the middle
    axis line style={<->}, % arrows on the axis
    xlabel={$x$},          % default put x on x-axis
    ylabel={$f(x)$},          % default put y on y-axis
    },
}

\usepackage{hyperref}

\usepackage[accepted]{icml2018}

\icmltitlerunning{Model-Based Value Estimation for Efficient Model-Free Reinforcement Learning}

\begin{document}

\twocolumn[
\icmltitle{Model-Based Value Expansion\\
for Efficient Model-Free Reinforcement Learning}

\icmlsetsymbol{equal}{*}

\begin{icmlauthorlist}
\icmlauthor{Vladimir Feinberg}{ucb}
\icmlauthor{Alvin Wan}{ucb}
\icmlauthor{Ion Stoica}{ucb}
\icmlauthor{Michael I.~Jordan}{ucb}
\icmlauthor{Joseph E.~Gonzalez}{ucb}
\icmlauthor{Sergey Levine}{ucb}
\end{icmlauthorlist}

\icmlaffiliation{ucb}{EECS Department, UC Berkeley, Berkeley, CA, USA}

\icmlcorrespondingauthor{Vladimir Feinberg}{vladf@berkeley.edu}

\icmlkeywords{Reinforcement Learning, Model-free, Model-based}

\vskip 0.3in
]

\printAffiliationsAndNotice{}  % leave blank if no need to mention equal contribution

\begin{abstract}

Recent model-free reinforcement learning algorithms have proposed incorporating learned dynamics models as a source of additional data with the intention of reducing sample complexity. 
Such methods hold the promise of incorporating imagined data coupled with a notion of model uncertainty to accelerate the learning of continuous control tasks.
Unfortunately, they rely on heuristics that limit usage of the dynamics model. 
% due to apparent over-use of imagined data. 
We present \emph{model-based value expansion}, which controls for uncertainty in the model by only allowing imagination to fixed depth. By enabling wider use of learned dynamics models within a model-free reinforcement learning algorithm, we improve value estimation, which, in turn, reduces the sample complexity of learning.

\end{abstract}
\section{Introduction}

% NAF - https://arxiv.org/abs/1603.00748
% Horia LQR - https://arxiv.org/abs/1710.01688
% SVG - https://arxiv.org/abs/1510.09142
% I2A - https://arxiv.org/abs/1707.06203
% Degris 2012 / Offpac - https://arxiv.org/abs/1205.4839
% VPN - https://arxiv.org/abs/1707.03497
% Kalweit / MA-BDDPG - http://proceedings.mlr.press/v78/kalweit17a/kalweit17a.pdf
% ME-TRPO - https://openreview.net/forum?id=SJJinbWRZ
% param noise - https://arxiv.org/abs/1706.01905
% Dyna - https://pdfs.semanticscholar.org/b5f8/a0858fb82ce0e50b55446577a70e40137aaf.pdf
% DPG - proceedings.mlr.press/v32/silver14.pdf

% We focus on a reinforcement learning (RL) setting with a fully-observable Markov decision process (MDP) with deterministic but unknown dynamics and known reward.
% %The environment determines the dynamics and reward of the process, which describe how transitions occur from initial states and actions selected by an agent to the resulting states and received rewards. The objective is to construct a policy, a mapping from states to rewards, which maximizes the discounted sum of rewards received over the agent's entire trajectory.
% While this general setting can be applied to a variety of tasks, we focus on continuous state and action spaces for robotic control.

\begin{wrapfigure}{r}{4cm}
\begin{center}
\includegraphics[width=0.5\columnwidth]{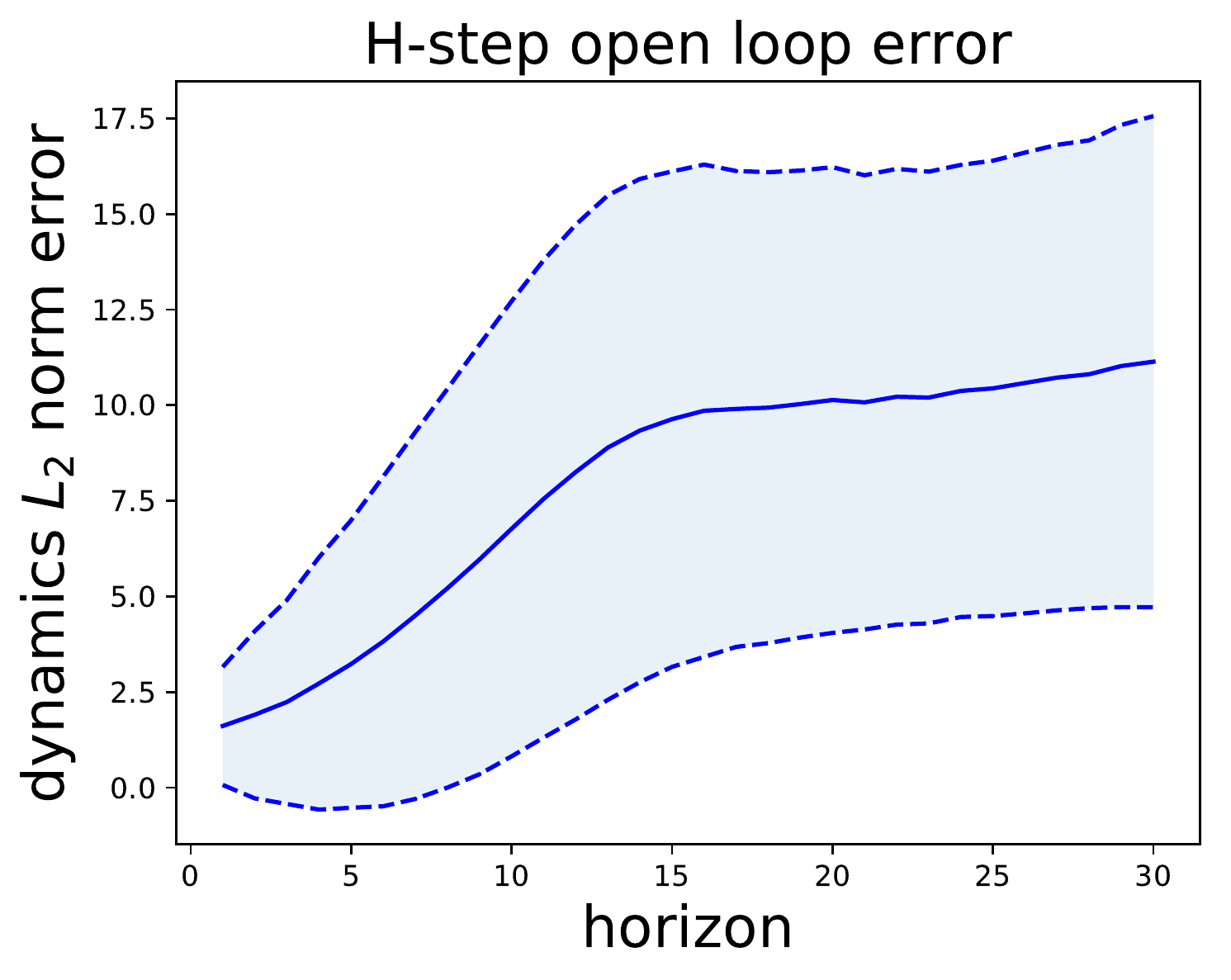}
\end{center}
\caption{The mean $L_2$ error, with standard deviation, for open-loop dynamics prediction of a \textit{cheetah} agent increases in the prediction horizon.}
\label{fig:dynmse}
\end{wrapfigure}

Recent progress in model-free (MF) reinforcement learning
% based around learning value functions 
has demonstrated the capacity of rich value function approximators to master complex tasks.
However, these model-free approaches require access to an impractically large number of training interactions for most real-world problems.
% sufficiently large volume of data. 
% Moreover implicit planning through value optimization has demonstrated strong asymptotic performance, but at the expense of requiring many samples from the environment. 
In contrast, model-based (MB) methods can quickly arrive at near-optimal control with learned models under fairly restricted dynamics classes \cite{lqr}. 
In settings with nonlinear dynamics, fundamental issues arise with the MB approach: \emph{complex dynamics demand high-capacity models, which in turn are prone to overfitting in precisely those low-data regimes where they are most needed.} 
Model inaccuracy is further exacerbated by the difficulty of long-term dynamics predictions (Fig.~\ref{fig:dynmse}).
%(see Fig.~\ref{fig:dynmse}).
% , resulting in unstable learning (see Fig.~\ref{fig:dynmse}). 
%While models can improve learning efficiency, planning against an inaccurate model can have a substantial adverse effect on the learned policy.
% tends to produce extremely poor performance.
%The lack of robustness guarantees in the nonlinear case leads to overfitting: even successfully planning to a incorrect view of the world results in failure.
%%SL.02.07: nice figure, but maybe a bit too technical for intro?
%% VF --> What makes it too technical? I though this could be a very motivating intro picture saying why we should focus on simulation depth as the fundamental measure of uncertainty
% \begin{figure}[!h]
% \end{figure}

The MF and MB approaches have distinct strengths and weaknesses: expressive value estimation MF methods can achieve good asymptotic performance but have poor sample complexity, while MB methods exhibit efficient learning but struggle on complex tasks. 
In this paper, we seek to \emph{reduce sample complexity while supporting complex non-linear dynamics by combining MB and MF learning techniques through disciplined model use for value estimation}.

We present model-based value expansion (MVE), a hybrid algorithm that uses a dynamics model to simulate the short-term horizon and Q-learning to estimate the long-term value beyond the simulation horizon.
% t for its capacity to simulate a fixed horizon into the future. 
This improves Q-learning by providing higher-quality target values for training.
Splitting value estimates into a near-future MB component and a distant-future MF component offers a model-based value estimate that (1) creates a decoupled interface between value estimation and model use and (2) does not require differentiable dynamics.

In this scheme, our trust in the model informs the selection of the horizon up to which we believe the model can make accurate estimates. 
% which provides an intuitive hyperparameter. 
This horizon is an interpretable metric applicable to many dynamics model classes. Prior approaches that combine model-based and model-free RL, such as model-based acceleration \cite{naf}, incorporate data from the model directly into the model-free RL algorithm, which we show can lead to poor results. Alternatively, imagination-augmented agents (I2A) offloads all uncertainty estimation and model use into an implicit neural network training process, inheriting the inefficiency of model-free methods \cite{i2a}.

By incorporating the model into Q-value target estimation, we only require the model to be able to make forward predictions. In contrast to stochastic value gradients (SVG), we make no differentiability assumptions on the underlying dynamics, which usually include non-differentiable phenomena such as contact interactions \cite{svg}. Using an approximate, few-step simulation of a reward-dense environment, the improved value estimate provides enough signal for faster learning for an actor-critic method (Fig.~\ref{fig:hc-true}). Our experimental results show that our method can outperform both fully model-free RL algorithms and prior approaches to combining real-world and model-based rollouts for accelerated learning.
The remainder of this paper is organized along the following contributions:
\begin{itemize}
\item A general method to reduce the value estimation error in algorithms with model-free critics (Sec.~\ref{sec:mve}).
\item An evaluation of the reduced sample complexity resulting from better value estimation (Sec.~\ref{sec:mve-perf}).
\item A characterization of the error that arises from value-based methods applied to imaginary data (Sec.~\ref{sec:critic-improve}).
\end{itemize}

\section{Background}

A deterministic Markov decision process (MDP) is characterized by a set of possible actions $\mcA$ and states $\mcS$. Its dynamics are deterministic and are described by a state transition function $f:\mcS\times \mcA\rightarrow\mcS$ and the bounded reward function $r:\mcS\times\mcA\rightarrow \R$.
In this work, we consider continuous state and action spaces. 
For a deterministic policy $\pi:\mcS\rightarrow\mcA$, the action-value function $Q^\pi(s_0,a_0)=\sum_{t=0}^\infty \gamma^t r_t$ gives the deterministic $\gamma$-discounted cumulative reward, where 
% $s_{t+1}=f^\pi(s_t)$, $r_t=r(s_t,a_t)$, 
% $f^\pi=f(\cdot,\pi(\cdot))$, 
$s_{t+1}=f^\pi(s_t) = f(s_t,\pi(s_t)) $, $a_t=\pi(s_t)$,
 and $r_t=r(s_t,a_t)$. We will use $\hat Q$ to refer to some estimate of $Q^\pi$.
Finally, recall that we may express the value function $V^\pi(s)=Q^\pi(s,\pi(s))$. We will denote state-value estimates with $\hat V$.
The policy $\pi$ is implicitly parameterized by $\theta$. 
We indicate this with subscripts $\pi_\theta$ when necessary.
% , but otherwise avoid the cumbersome notation.

The objective is to maximize $J_{d_0}(\pi)=\E_{d_0}\ha{V^\pi(S)}$, where $S\sim d_0$ is sampled from the initial state distribution. With access to only off-policy states from an exploration policy, summarized by some empirical distribution $\nu$, we consider the proxy $J_\nu$ instead, as is typical \cite{degris}.

\subsection{Deterministic Policy Gradients (DPG)}

\citet{dpg} describe an off-policy actor-critic MF algorithm. For a deterministic, parameterized policy $\pi_\theta$ and parameterized critic $\hat Q$ (we leave its parameters implicit), \citet{dpg} prove that a tractable form of the policy gradient $\nabla_\theta J_{d_0}(\theta)$ exists when $f$ is continuously differentiable with respect to actions taken by the policy. This result, the deterministic policy gradient theorem, expresses the policy gradient as an expectation over on-policy data.

To encourage exploration, the data collected may be off-policy. To reconcile the use of off-policy data with an on-policy estimator, \citet{dpg} briefly note an analogous off-policy policy improvement theorem in the case of finite state spaces \cite{degris}.

\subsection{Continuous DPG}

% We found that 
In the continuous case, the conditions under which the off-policy policy improvement theorem holds are distinct from the discrete case. We emphasize these conditions here and provide a proof for the statement in \citet{dpg} inspired by the discrete case \citep{degris}.

\begin{thm}[\textbf{Off-Policy Deterministic Continuous Policy Improvement Theorem}] 
% \cite{degris}. 
\label{thm:offpolicy_improvement}
Let $\beta$ be an off-policy distribution of states and set
\begin{equation}
g_\theta(s)=D_{\pi(s)}(\theta)^\top\nabla_a Q^{\pi_\theta}(s, a)\big|_{a=\pi(s)},
\label{eq:ascentdir}
\end{equation}
with $D_{\pi(s)}(\theta)$ the Jacobian of $\theta\mapsto \pi_{\theta}(s)$. Then $\E_\beta g_\theta(S)$ ascends $J_\beta(\theta)$ if the following conditions hold $\beta$-almost always: 
(1) $Q^{\pi_\theta}$ must be differentiable with respect to the action at $(s, \pi_\theta(s))$, (2) the Jacobian  $D_{\pi(s)}(\theta)$ must exist, and (3) $g_\theta(s)$ is nonzero.
\end{thm}

\textit{Proof}. See the appendix (Sec.~\ref{sec:dpg-continuous}).

Critically, we do not require DPG's assumptions of continuous differentiability in the reward or dynamics functions. This will allow model-based value expansion to be both theoretically and practically compatible with arbitrary forward models, including discrete event simulators, and settings with non-differentiable physics. 

Finally, a proxy $\hat Q$ may be used as a replacement for the true $Q^\pi$ term to approximate the ascent direction $g_\theta$ in \eqnref{eq:ascentdir}, as described in deep deterministic policy gradients (DDPG) \cite{ddpg}.

\section{Model-Based Value Expansion}\label{sec:mve}

MVE improves value estimates for a policy $\pi$ by assuming we an approximate dynamical model $\hat f:\mcS\times\mcA\rightarrow\mcS$ and the true reward function $r$. Such an improved value estimate can be used in training a critic for faster task mastery in reward-dense environments (\figref{fig:hc-true}). We assume that the model is accurate to depth $H$; that is, for a fixed policy $\pi$, we may use $\hat f$ to evaluate (``imagine") future transitions that occur when taking actions according to $\pi$ with $\hat f^\pi=\hat f(\cdot,  \pi(\cdot))$. We use these future transitions to estimate value.

\begin{dfn}[\textbf{$H$-Step Model Value Expansion}]\label{def:mve}
Using the imagined rewards reward $\hat r_t=r(\hat s_t,\pi(\hat s_t))$ obtained from our model $\hat s_t=\hat f^\pi(\hat s_{t-1})$ under $\pi$
we define the $H$-step model value expansion (MVE) estimate for the value of a given state $V^\pi(s_0)$: % in \eqnref{eq:mve},
% relying on an existing critic $\hat V$ and dynamics model $\hat f$ by setting
\begin{equation}
\hat V_H(s_0)=\sum_{t=0}^{H-1}\gamma^t\hat r_t + \gamma^H\hat V(\hat s_H)\label{eq:mve}\,.
\end{equation}
\end{dfn}

The $H$-step model value expansion in \dfnref{def:mve} 
% \eqnref{eq:mve} 
decomposes the state-value estimate at $s_0$ into the component predicted by learned dynamics $\sum_{t=0}^{H-1}\gamma^t\hat  r_t$ and the tail, estimated by $\hat V$. This approach can be extended to stochastic policies and dynamics by integrating Eq.~\ref{eq:mve} with a Monte Carlo method, assuming a generative model for the stochastic dynamics and policy. Since $\hat r_t$ is derived from actions $\hat a_t=\pi(\hat s_t)$, this is an on-policy use of the model and, even in the stochastic case, MVE would not require importance weights, as opposed to the case of using traces generated by off-policy trajectories.

While MVE is most useful in settings where the $H$ step horizon is not sparse, even in sparse reward settings predicting the future state will improve critic accuracy.
Finally, MVE may be applied to state-action estimates: in this case $\hat a_0\triangleq a_0$ while $\hat a_t=\pi(\hat s_t)$ for $t>0$ and $\hat V(\hat s_H)\triangleq \hat Q(\hat s_H,\hat a_H)$ and may be used for estimating $\hat Q_H(s_0,a_0)$ with a state-action critic $\hat Q$.

\subsection{Value Estimation Error}\label{sec:value-error}

In this section, we discuss the conditions under which the MVE estimate improves the mean-squared error (MSE):
\begin{equation}
    \MSE_\nu(V)=\mathExpUnder_{S\sim\nu}\ha{\pa{V(S)-V^\pi(S)}^2}\label{eq:mse},
\end{equation}
of the original critic $\hat V$ with respect to $\nu$, a distribution of states.
We emphasize that, even assuming an ideal model, the mere combination of MB and MF does not guarantee improved estimates. Further, while our analysis will be conducted on state-value estimation, it may be naturally extended to state-action-value estimation.

Recall the $H$-depth model accuracy assumption: if we observe a state $s_0$ we may imagine $\hat s_t\approx s_t$, so $\hat a_t\approx a_t$ and $\hat r_t\approx r_t$ for $t\le H$. If the modelling assumption holds for some fixed $s_0$, we have
$$\hat V_H(s_0)-V^{\pi}(s_0)\approx \gamma^H\pa{\hat V( s_H)-V^{\pi}(s_H)}\,.$$
This translates to an expression for MVE MSE,
$$
\MSE_\nu(\hat V_H)\approx \gamma^{2H}\MSE_{(f^\pi)^H\nu} (\hat V)\,,
$$
where $(f^\pi)^H\nu$ denotes the pushforward measure resulting from playing $\pi$ $H$ times starting from states in $\nu$. This informal presentation demonstrates the relation of MVE MSE to the underlying critic MSE by assuming the model is nearly ideal. We verify that the informal reasoning above is sound in the presence of model error.

\begin{thm}[\textbf{Model-Based Value Expansion Error}] 
Define $s_t,a_t,r_t$ to be the states, actions, and rewards 
resulting from following policy $\pi$ using the true dynamics $f$ starting at $s_0\sim\nu$ and
% that come about from acting according to $\pi:\mcS\rightarrow\mcA$ in the true environment with dynamics $f$. 
analogously define $\hat s_t,\hat a_t,\hat r_t$ using the learned dynamics $\hat f$ in place of $f$.
Let the reward function $r$ be $L_r$-Lipschitz and the value function $V^\pi$ be $L_V$-Lipschitz. 
% $H$-step roll-out.
% be constructed from starting at $s_0$ and using the learned dynamics $\hat f$ to carry out an $H$-step roll-out. 
Let $\epsilon$ be a be an upper bound
$$
\max_{t\in[H]}{\E\left[\norm{\hat s_t- s_t}^2\right]}\le \epsilon^2,
$$
% If the model has risk $\epsilon$ at depth $H$, i.e., $\max_{t\in[H]}{\E\norm{\hat s_t- s_t}^2}\le \epsilon^2$, 
% and $r,V^\pi$ are $L_r,L_V$-Lipschitz functions of state, 
on the model risk for an $H$-step rollout.  Then
$$
\MSE_\nu(\hat V_H)\le c_1^2\epsilon^2+ (1+c_2\epsilon)\gamma^{2H}\MSE_{(\hat f^\pi)^H\nu} (\hat V)\,,
$$
where $c_1,c_2$ grow at most linearly in $L_r,L_V$ and are independent of $H$ for $\gamma <1$. We assume $\MSE_{(\hat f^\pi)^H\nu} (\hat V)\ge \epsilon^2$ for simplicity of presentation, but an analogous result holds when the critic outperforms the model.
\end{thm}

\textit{Proof}. See the appendix (Sec.~\ref{sec:mve-error}).

Sufficient conditions for improving on the original critic $\MSE_\nu (\hat V)$ are then small $\epsilon$, $\gamma < 1$, and the critic being at least as accurate on imagined states as on those sampled from $\nu$,
\begin{equation}
\MSE_\nu (\hat V)\ge \MSE_{(\hat f^\pi)^H\nu} (\hat V).\label{eq:offdist}
\end{equation}
However, if $\nu$ is an arbitrary sampling distribution, such as the one generated by an exploratory policy, the inequality of Eq.~\eqref{eq:offdist} rarely holds. In particular, this naive choice results in poor performance overall (Fig.~\ref{fig:ablation}) and counteracts the benefit of model-based reward estimates, even assuming a perfect oracle dynamics model (Fig.~\ref{fig:hc-true}). Thus, if $\hat V$ is trained on Bellman error from $\nu$, then the \textbf{distribution mismatch} between $(f^\pi)^H\nu$ and $\nu$ eclipses the benefit of $\gamma^{2H}$. In effect, any model-based approach evaluating the critic $\hat V$ on imaginary states from $(f^\pi)^H \nu$ must be wary of only training its critic on the real distribution $\nu$. We believe this insight can be incorporated into a variety of works similar to MVE, such as value prediction networks \cite{vpn}.

We propose a solution to the distribution-mismatch problem by observing that the issue disappears if $(f^\pi)^H\nu=\nu$, i.e., the training distribution $\nu$ is a fixed point of $f^\pi$. In practice, given an arbitrary off-policy distribution of state-action pairs $\beta$ we set $\nu=\E\ha{(f^\pi)^T\beta}$ as an approximation to the fixed point, where $T\sim\Uniform\ca{0, \cdots, H-1}$. If we then sample a state $\hat s_T|T\sim \pa{f^\pi}^T\beta$, our model accuracy assumption dictates that we may accurately simulate states $\ca{\hat  s_{T+i}}_{i=1}^{H-T}$ arising from playing $\pi$ starting at $\hat{s}_T$. These simulated states can be used to construct $k$-step MVE targets $\hat V_{k}(\hat s_T)$ accurately while adhering to assumptions about the model by setting $k=H-T$. These targets can then be used to train $\hat V$ on the entire support of $\nu$, instead of just $\beta$.

Since we do not have access to the true MSE of $\hat V$, we minimize its Bellman error with respect to $\nu$, using this error as a proxy. In this context, using a target $\hat V_k(\hat s_T)$ is equivalent to training $\hat V$ with imagined TD-$k$ error. This \textbf{TD-$k$ trick} enables us to skirt the distribution mismatch problem to the extent that $\nu$ is an approximate fixed point. We find that the TD-$k$ trick greatly improves task performance relative to training the critic on $\beta$ alone (Fig.~\ref{fig:ablation}).

\subsection{Deep Reinforcement Learning Implementation}

In the preceding section, we presented an analysis that motivates our model-based value expansion approach. 
In this section, we will present a practical implementation of this approach for high-dimensional continuous deep reinforcement learning.  
We demonstrate how to apply MVE in a general actor-critic setting
to improve the target $Q$-values, with the intention of achieving faster convergence. Our implementation relies on a parameterized actor $\pi_\theta$ and critic $Q_\varphi$, but note that the separate parameterized actor may be removed if it is feasible to compute $\pi(s)=\argmax_a  Q_\varphi(s,a)$.

We assume that the actor critic method supplies a differentiable actor loss $\ell_{\mathrm{actor}}$ and a critic loss $\ell_{\mathrm{critic}}^{\pi,Q}$. These losses are functions of $\theta,\varphi$ as well as transitions $\tau=(S,A,R,S')$ sampled from some distribution $\mcD$. For instance, in DDPG, the $\theta$ gradient of $\E_\mcD\ha{\ell_{\mathrm{actor}}(\theta,\varphi, \tau)}=\E_\mcD\ha{Q_\varphi(S, \pi_\theta(S)}$ approximately ascends $J_{\mcD}$ per the deterministic continuous policy improvement theorem \cite{ddpg}. The DDPG critic loss depends on a target actor $\pi$ and critic $Q$: $\ell_{\mathrm{critic}}^{\pi,Q}(\varphi,\tau)=\pa{Q_\varphi(S,A)-\pa{R+\gamma Q(S', \pi(S'))}}^2$.

MVE relies on our approximate fixed point construction from an empirical distribution of transitions $\beta$. Recall our approximation from the previous section, which relies on the current policy to imagine up to $H$ steps ahead: $\nu(\theta',\hat f)=\frac{1}{H}\sum_{t=0}^{H-1}(\hat f^{\pi_{\theta'}})^t\beta$, where $\hat f^{\pi_{\theta'}}(\tau)=(S', A', r(S', A'), \hat f(S', A'))$ and $A'=\pi_{\theta'}(S')$. Thus, sampling transitions from $\nu$ is equivalent to sampling from any point up to $H$ imagined steps into the future, when starting from a state sampled from $\beta$. The MVE-augmented method follows the usual actor-critic template, but critic training uses MVE targets and transitions sampled from $\nu$ (Alg.~\ref{alg:mve}).

We imagine rollouts with the target actor, whose parameters are exponentially weighted averages of previous iterates, for parity with DDPG, which uses a target actor to compute target value estimates. Taking $H=0$, $\nu(\theta', \hat f)=\beta$, we recover the original actor-critic algorithm. Our implementation uses multi-layer fully-connected neural networks to represent both the $Q$-function and the policy. We use the actor and critic losses described by DDPG \cite{ddpg}.

\begin{algorithm}[!ht]
  \caption{Use model-based value expansion to enhance critic target values in a generic actor-critic method abstracted by $\ell_{\mathrm{actor}},\ell_{\mathrm{critic}}$. Parameterize $\pi,Q$ with $\theta,\varphi$, respectively. We assume $\hat f$ is selected from some class of dynamics models and the space $\mcS$ is equipped with a norm.} \label{alg:mve}
\begin{algorithmic}[1]
  \Procedure{MVE-AC}{initial $\theta,\varphi$}
  \State Initialize targets $\theta'=\theta,\varphi'=\varphi$
  \State Initialize the replay buffer $\beta\gets \emptyset$
  \While{not tired}
    \State Collect transitions from any exploratory policy
    \State Add observed transitions to $\beta$
    \State Fit the dynamics
    $$\hat f\gets\argmin_{f}\mathExpUnder_\beta\ha{\norm{f(S,A)-S'}^2}$$
    \For{a fixed number of iterations}
        \State sample $\tau_0\sim\beta$
        \State update $\theta$ with $\nabla_\theta \ell_{\mathrm{actor}}\pa{\pi_\theta,Q_\varphi,\tau_0}$
        \State imagine future transitions for $t\in [H-1]$
        $$
        \tau_{t}=\hat f^{\pi_{\theta'}}(\tau_{t-1})
        $$\label{line:sample}\vspace{-\baselineskip}
        \State $\forall k$ define $\hat Q_k$ as the $k$-step MVE of $Q_{\varphi'}$
        \State update $\varphi$ with $\nabla_\varphi\sum_t\ell_{\mathrm{critic}}^{\pi_{\theta'},\hat Q_{H-t}}\pa{\varphi,\tau_t}/H$ \label{line:critic-update}\vspace{-\baselineskip}
        \State update targets $\theta',\varphi'$ with some decay
    \EndFor
  \EndWhile
\State \Return $\theta,\varphi$ 
\EndProcedure
\end{algorithmic}
\end{algorithm}

Importantly, we do not use an imagination buffer to save simulated states, and instead generate simulated states on-the-fly by sampling from $\nu(\theta',\hat f)$. We perform a stratified sampling from $\nu(\theta',\hat f)$, with $H$ dependent samples at a time, for each $t\in \{0,\cdots,H-1\}$ in Line~\ref{line:sample}. First, we sample a real transition $\tau_0=(s_{-1},a_{-1},r_{-1},s_0)$ from $\beta$, the empirical distribution of transitions observed from interacting with the environment according to an exploratory policy. We use the learned dynamics model $\hat f$ to generate $\hat s_t$ and $\hat r_t$. Since $\pi_{\theta'}$ changes during the joint optimization of $\theta,\varphi$, these simulated states are discarded immediately after the batch. We then take a stochastic $\nabla_\varphi$ step to minimize $\nu$-based Bellman error of $Q_\varphi$,
$$
\frac{1}{H}\sum_{t=-1}^{H-1}\pa{ Q_\varphi (\hat s_t, \hat a_t)-
\pa{\sum_{k=t}^{H-1}\gamma^{k-t}\hat r_k+\gamma^H  Q_{\varphi'}\pa{\hat s_H,\hat a_H}}}^2\,,
$$
where $Q_{\varphi'}$ and and $\hat a_t=\pi_{\theta'}(\hat s_t)$ use target parameter values (Lines~\ref{line:sample}-\ref{line:critic-update} of Alg.~\ref{alg:mve}). As such, every observation of the Bellman error always relies on some real data.

For the dynamics $\hat f$, we use a neural network network with 8 layers of 128 neurons each with a fixed $10^{-3}$ learning rate trained to predict the difference in real-vector-valued states, similar to previous work \cite{metrpo}. While we expect more accurate and carefully tuned models to allow us to use large $H$, even a weak model with shared hyperparameters across all tasks from a flexible class suffices to demonstrate our point.

\section{Results}

We evaluate MVE on several continuous control environments. Experimental details are in Sec.~\ref{sec:exp-det}. We would like to verify the following:
\begin{itemize}
    \item Does MVE improve estimates of $Q^\pi$?
    \item Do the improved estimates result in faster mastery?
    \item Does the TD-$k$ trick resolve distribution mismatch?
\end{itemize}

In all of our experiments, we tune the baseline, DDPG, and report its best performance. For exploration, we use parameter-space noise \cite{paramnoise}. Every experiment and each setting uses the same adaptive parameter-space noise standard deviation target, so exploration is controlled to be the same in all trials. We then add the MVE extension (we do not tune DDPG parameters to MVE performance). To evaluate the effect of the TD-$k$, we evaluate against the naive approach of using $H$-step MVE $\hat Q$ estimates for Bellman error on states sampled from $\beta$. This is equivalent to using only the first term for $t=0$ in the sum of Line~\ref{line:critic-update} of Alg.~\ref{alg:mve}, i.e., updating critic parameters $\varphi$ with the gradient $\nabla_\varphi\ell_{\mathrm{critic}}^{\pi_{\theta'},\hat Q_H}(\varphi, \tau_0)$. Without the TD-$k$ trick the model is still used to simulate to depth $H$, but the opportunity to learn on a distribution of additional support is neglected.

We plot the mean learning curve, along with the standard deviation. We smooth each graph with a 20 point window (evaluation data is collected at intervals of at most $10^3$ timesteps).

\begin{figure}
\begin{center}
  \subfigure[]{\includegraphics[width=0.3\columnwidth]{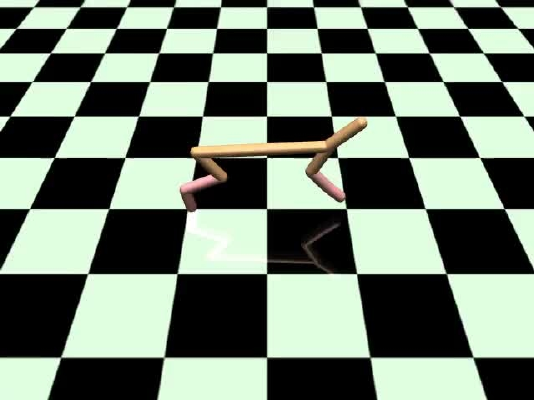}}
  \subfigure[]{\includegraphics[width=0.3\columnwidth]{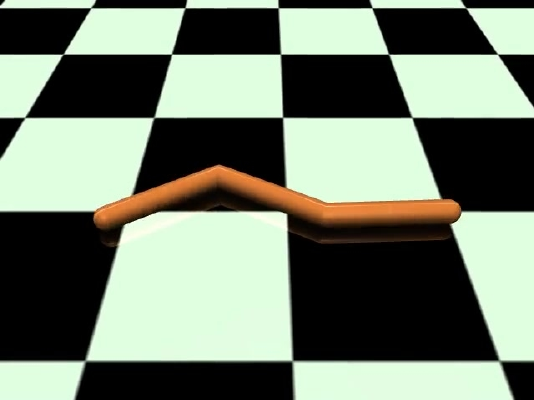}}
  \subfigure[]{\includegraphics[width=0.3\columnwidth]{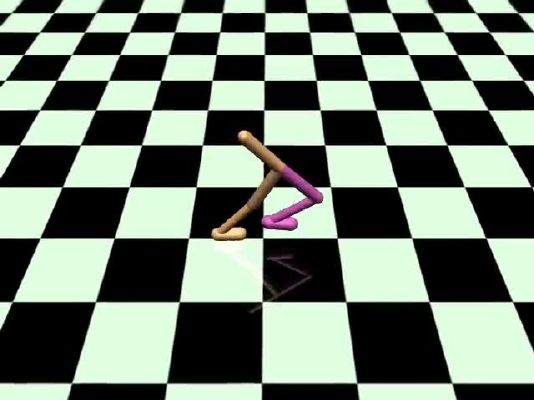}}
\end{center}
\caption{We use fully-observable analogues of typical Gym environments \cite{gym}. By default, the environments excerpt certain observed dimensions, such as the abscissa of (a), which are requisite for reward calculation. This is done as a supervised imposition of policy invariance to certain dimensions, which we remove for full observability. (a) shows \textit{Fully Observable Half Cheetah} (\textit{cheetah}); (b), \textit{Fully Observable Swimmer} (\textit{swimmer}); (c), \textit{Fully Observable Walker, 2-D} (\textit{walker}). Agents are rewarded for forward motion. We detail our changes in the Appendix (Sec.~\ref{sec:exp-det}).
}
\label{fig:envpics}
\end{figure}

\subsection{Performance}\label{sec:mve-perf}

\begin{figure*}[!h]
\begin{center}
\includegraphics[width=1.75\columnwidth]{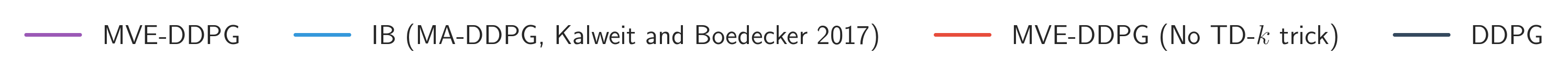}
\end{center}
\vspace{-2em}
\begin{center}
  \subfigure[]{\includegraphics[width=0.65\columnwidth]{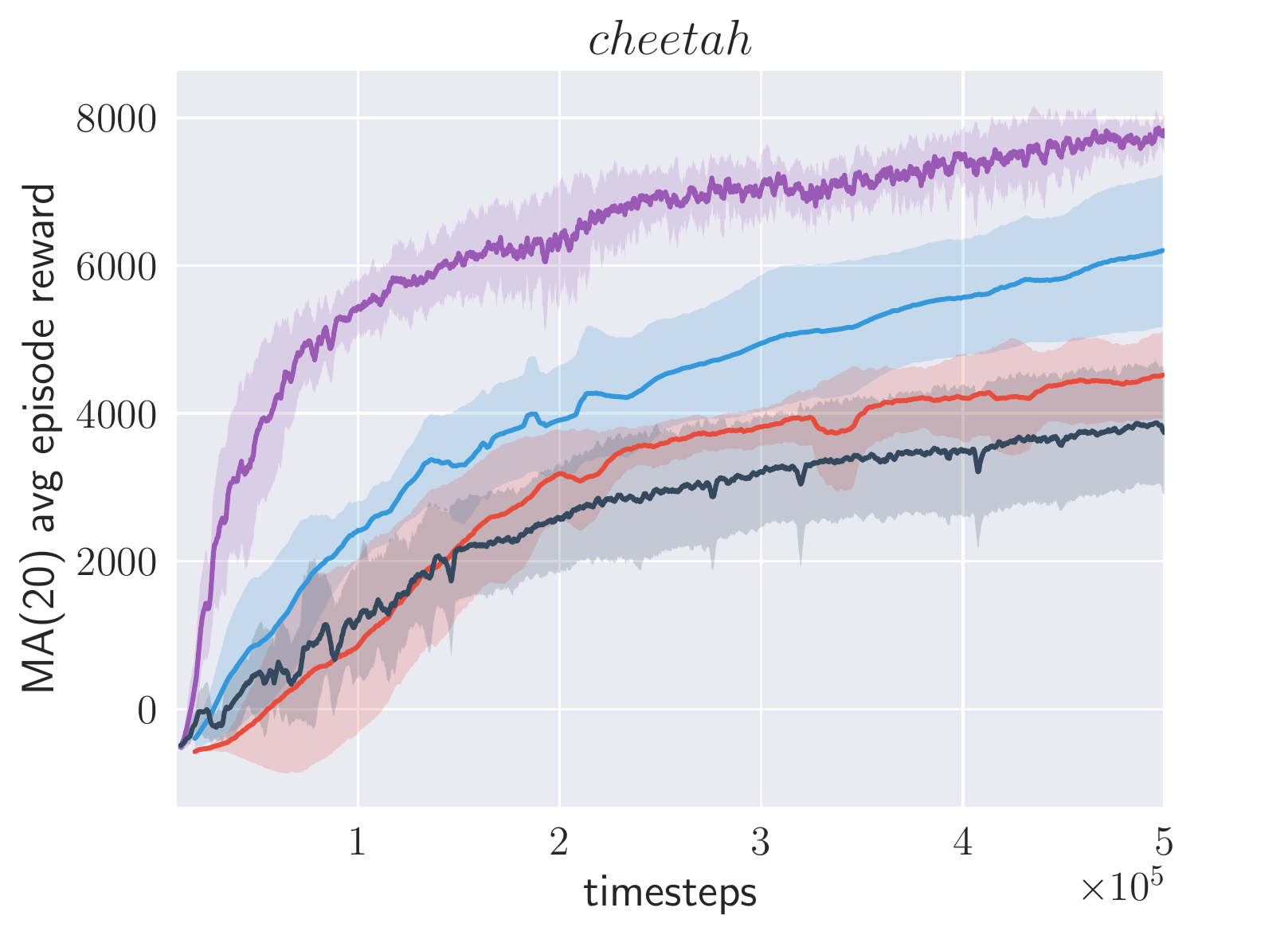}}
  \subfigure[]{\includegraphics[width=0.65\columnwidth]{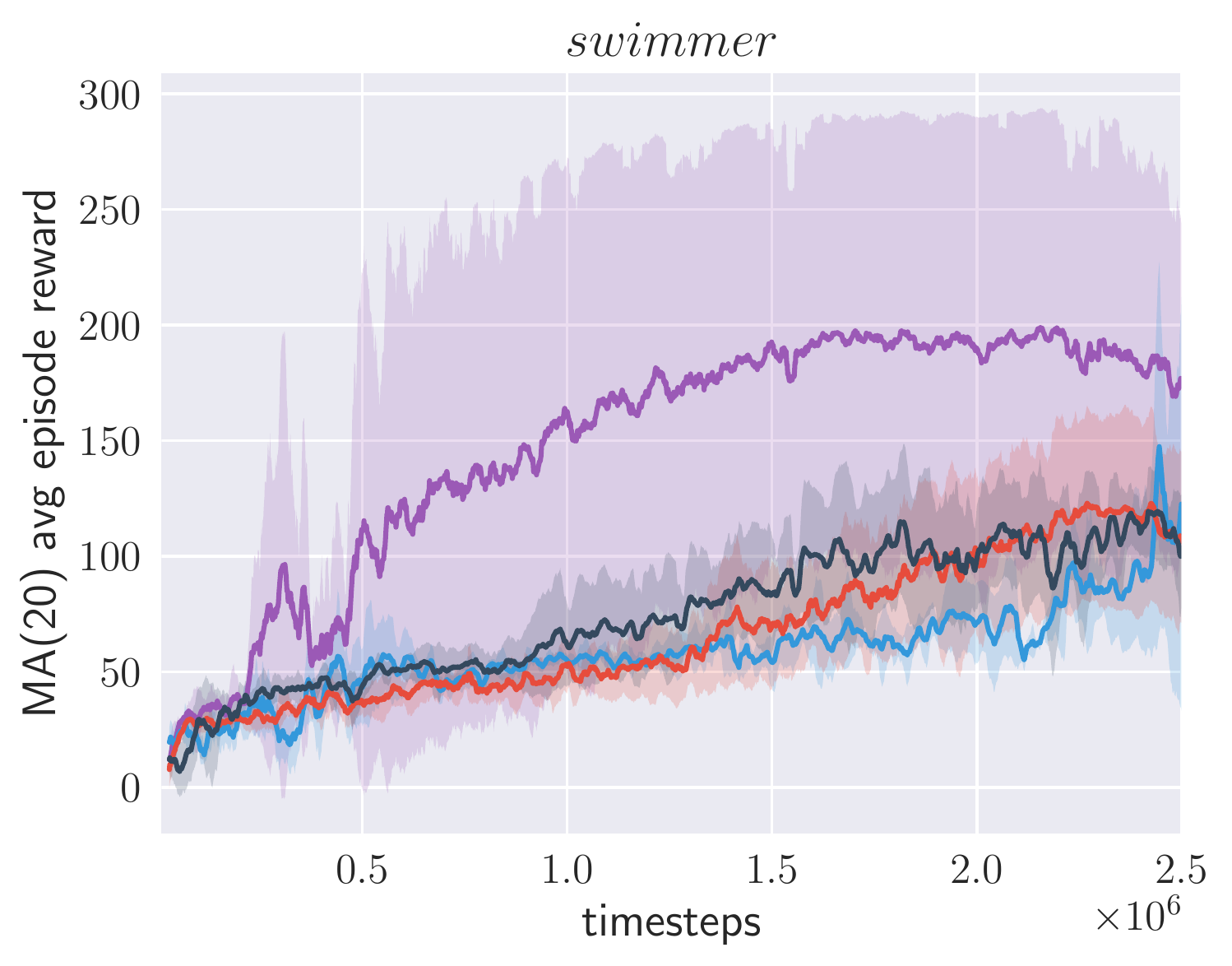}}
  \subfigure[]{\includegraphics[width=0.65\columnwidth]{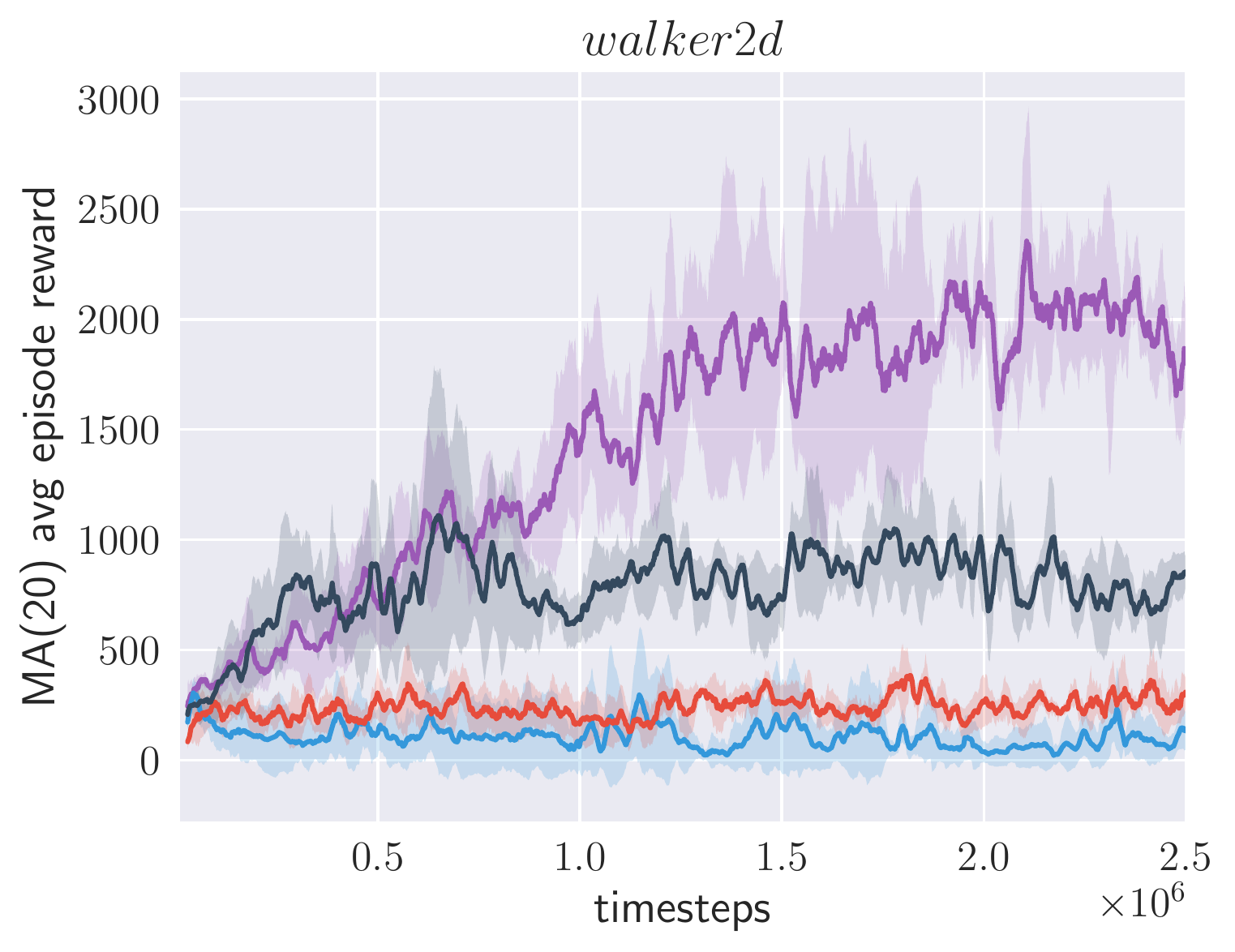}}
\end{center}

\caption{Learning curves comparing MVE with learned dynamics ({\color{purple}purple}), MVE without the TD-$k$ trick ({\color{orange}orange}), IB \citep{kalweit} ({\color{blue}blue}), and DDPG ({\color{black}black}) on (a) \textit{cheetah}, (b) \textit{swimmer}, and (b) \textit{walker}. We used $H=10$ for (a,b), but found the same dynamics class inadequate for walker, reducing \textit{walker} experiments to $H=3$ reduces the improvement MVE has to offer over DDPG, but it still exhibits greater robustness to the poor model fit than IB. Note that we use MA-DDPG, not MA-BDDPG in the IB approach. The bootstrap estimation in MA-BDDPG may reduce model use in some cases, so it is possible that MA-BDDPG would have improved performance in \textit{walker}, where the learned dynamics are poor compared to the other environments.}
\label{fig:ablation}
\end{figure*}

First, we evaluate that MVE-DDPG (with the TD-$k$ trick) improves in terms of raw reward performance by comparing its learning curves to those of the original DDPG, MVE-DDPG without the TD-$k$ trick, and imagination buffer (IB) approaches \cite{kalweit}. We find that MVE-DDPG outperforms the alternatives (Fig.~\ref{fig:ablation}).\footnote{We run IB with an imaginary-to-real ratio of 4, as used for 2 of 3 environments in \cite{kalweit}. We tested lower ratios on \textit{cheetah} but found they performed worse. Note that for parity with DDPG we ran MVE-DDPG with only 4 gradient steps.} Our result shows that incorporating synthetic samples from a learned model can drastically improve the performance of model-free RL, greatly reducing the number of samples required to attain good performance. As illustrated by the comparison to the IB baseline, this improvement is obtained only when carefully incorporating this synthetic experience via a short horizon and the TD-$k$ trick. As we will discuss in the next section, the specific design decisions here are critical for good results, which helps to explain the lack of success with learned neural network models observed with related methods in prior work~\cite{naf}.

MVE-DDPG improves on similar approaches, such as MA-DDPG, by its treatment of synthetic data obtained from the dynamics model. This alternative approach adds simulated data back into a separate imagination buffer, effectively modifying $\beta$ from Alg.~\ref{alg:mve} into a mixture between $\beta$ and simulated data (where the mixture is dependent on the relative number of samples taken from each buffer). This is problematic because the policy changes during training, so the data from this mixed distribution of real and fake data is stale relative to $\nu$ in terms of representing actions that would be taken by $\pi$. In our implementation of MA-DDPG, we do not reuse imagined states in this manner, but MVE-DDPG still outperforms MA-DDPG. We suspect this is due to two factors: (1) the staleness of the imaginary states in the IB approach, and (2) the delicate interaction between using more imaginary data and overtraining the actor. In order to use the additional synthetic data, an IB approach must take more gradient steps with imaginary batches. On the other hand, since MVE-DDPG uses a gradient averaged over both real and simulated data, the choice to make additional gradient steps becomes an independent consideration dependent on the stability of the actor-critic method being trained.

\subsection{MVE as Critic Improvement}\label{sec:critic-improve}

\begin{figure}[!h]
\begin{center}
\includegraphics[width=\columnwidth]{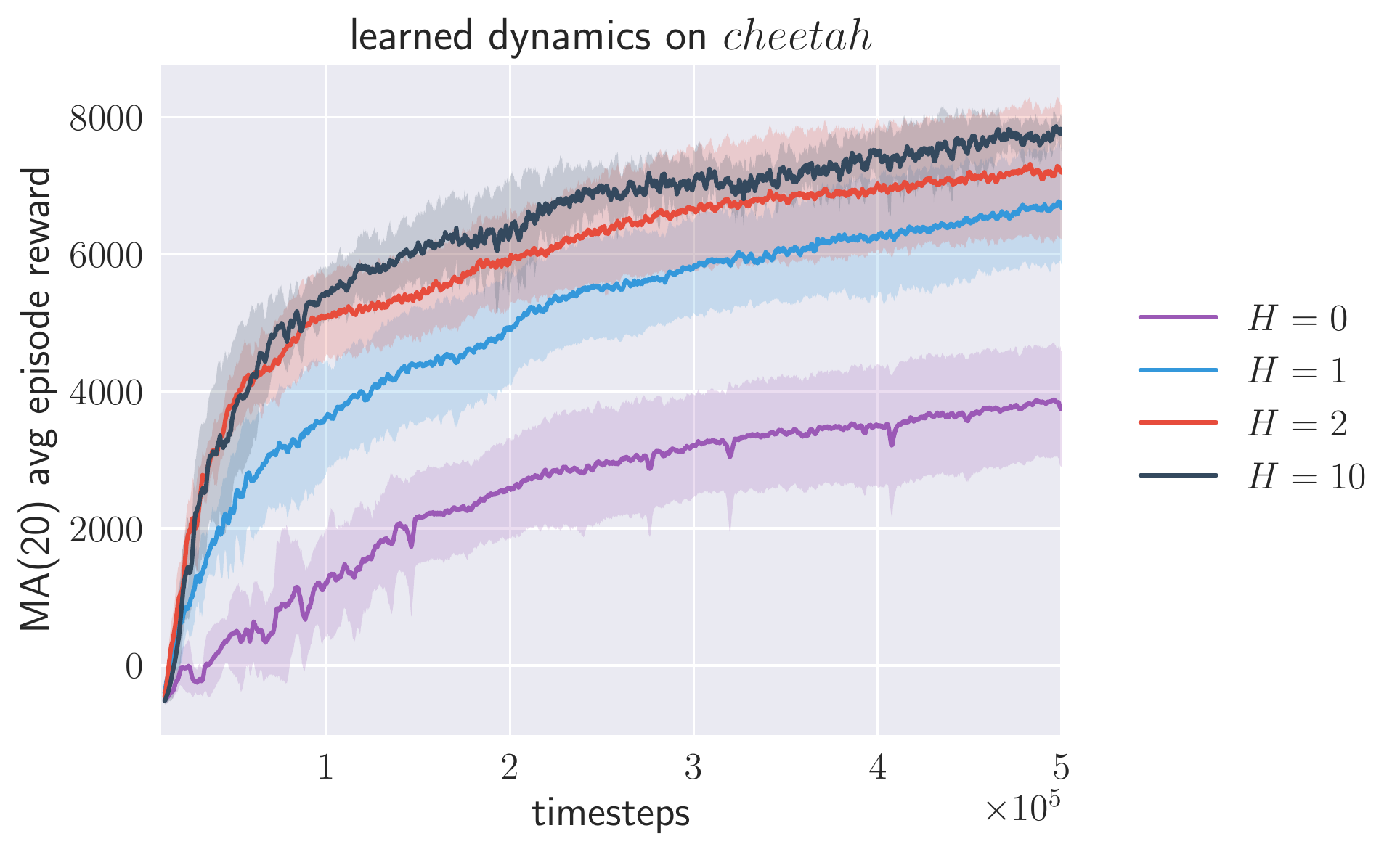}
\end{center}
\caption{Learning curves from \textit{cheetah} for MVE-DDPG \textit{with learned dynamics} at different model horizons $H$.
}
\label{fig:learned-hc-hs}
\end{figure}

\begin{figure}[!h]
\begin{center}
{\includegraphics[width=0.43\columnwidth]{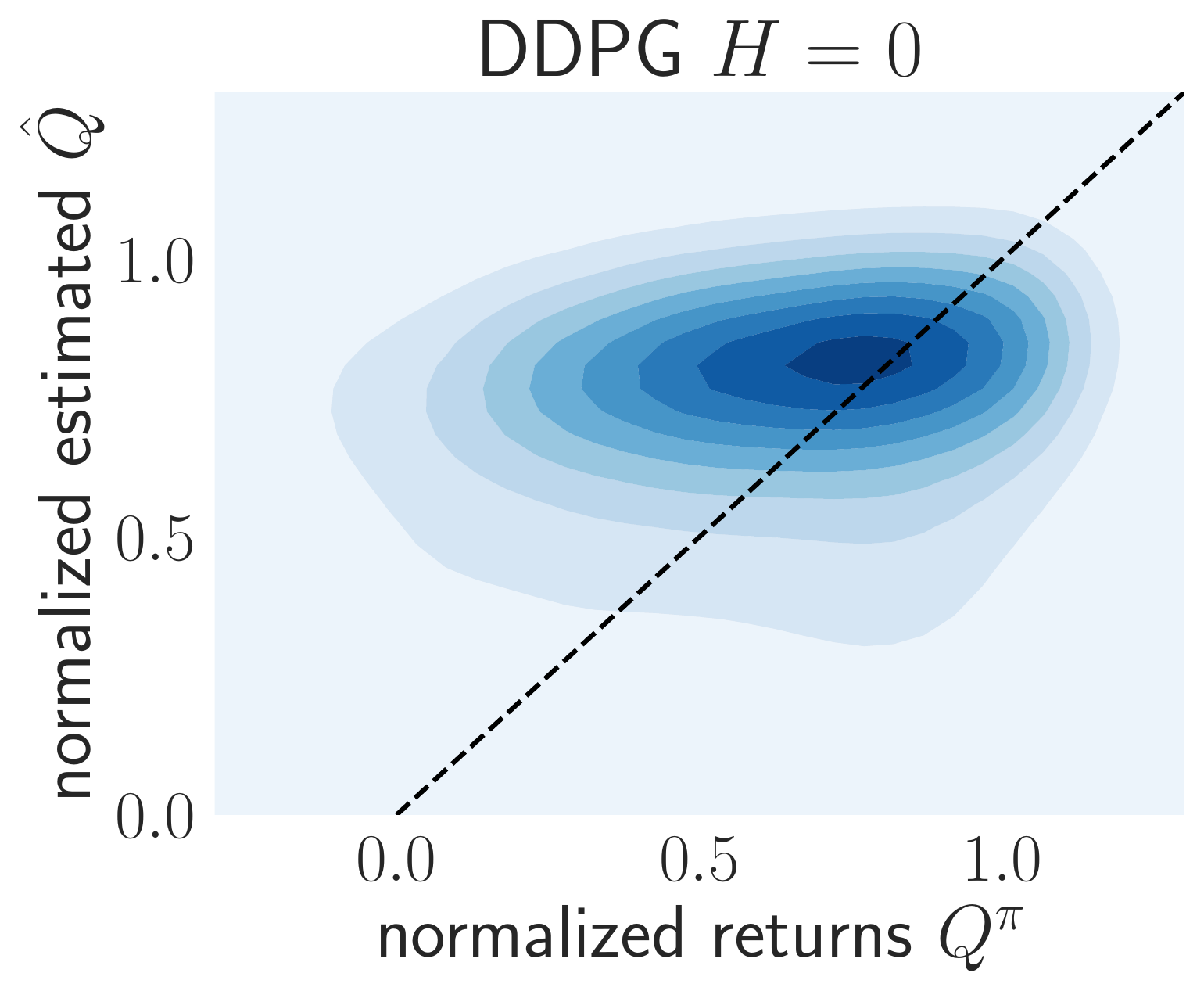}}
{\includegraphics[width=0.47\columnwidth]{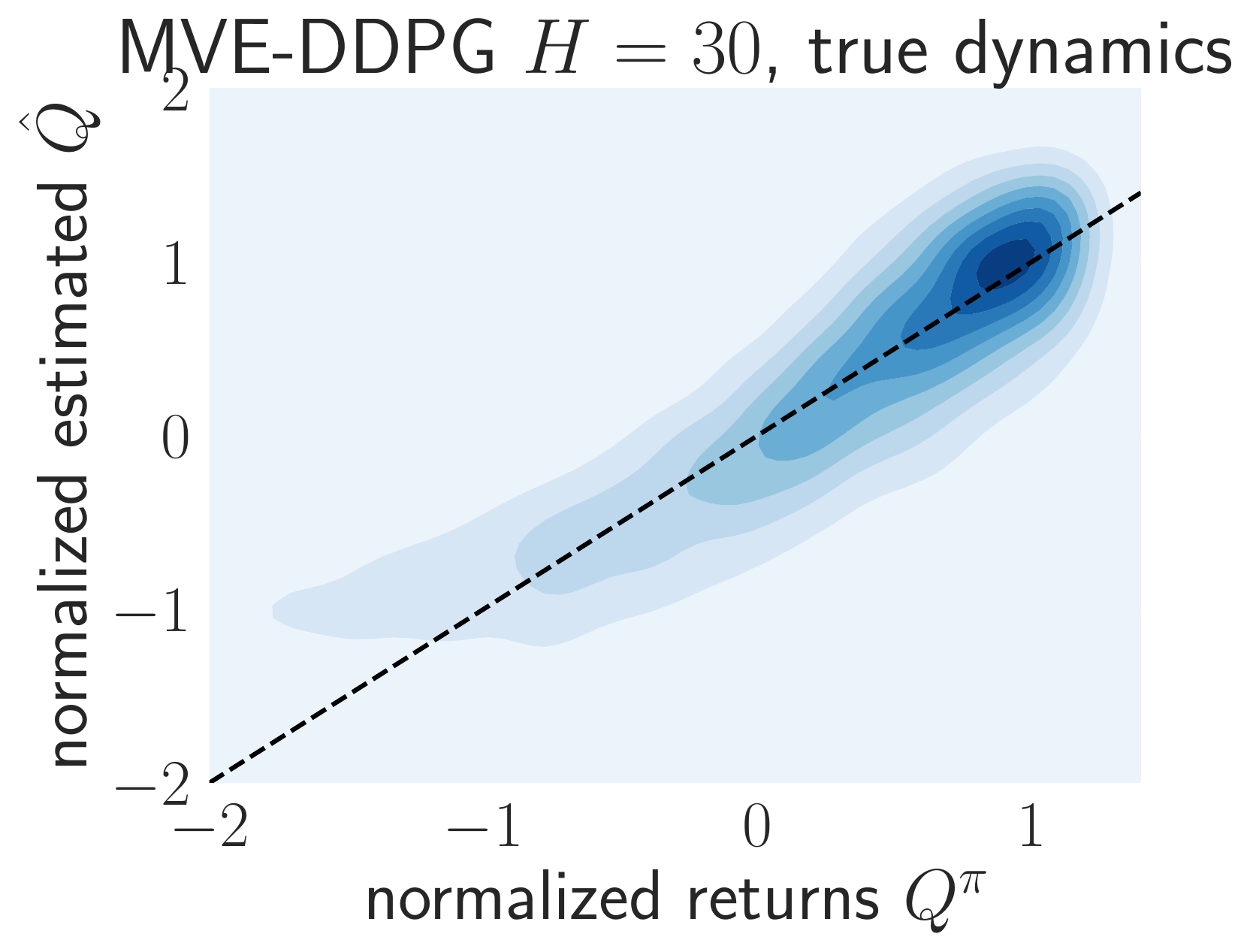}}
\end{center}
\caption{We plot the true observed cumulative discounted returns against those predicted by the critic for \textit{cheetah} at the end of training (both values are normalized), reconstructing Fig.~3 of \citep{ddpg}. The dotted black line represents unity. An ideal critic concentrates over the line. We verify that with MVE at $H=30$ with the true dynamics model trains a critic with improved $Q$ values relative to the original DDPG algorithm. Both of the above runs use a reduced mini-batch size because oracle dynamics are expensive to compute.
}
\label{fig:qdensity}
\end{figure}

We make sure that MVE can make use of accurate models to improve the critic value estimate (Fig.~\ref{fig:learned-hc-hs}). The improved critic performance results in faster training compared to the $H=0$ DDPG baseline on \textit{cheetah}. Also, we replicate the density plot from DDPG to analyze $Q$ accuracy directly, from which it is clear that MVE improves the critic by providing better target values (Fig.~\ref{fig:qdensity}).

\begin{figure*}[!h]
\begin{center}
\subfigure[]{\includegraphics[width=0.9\columnwidth]{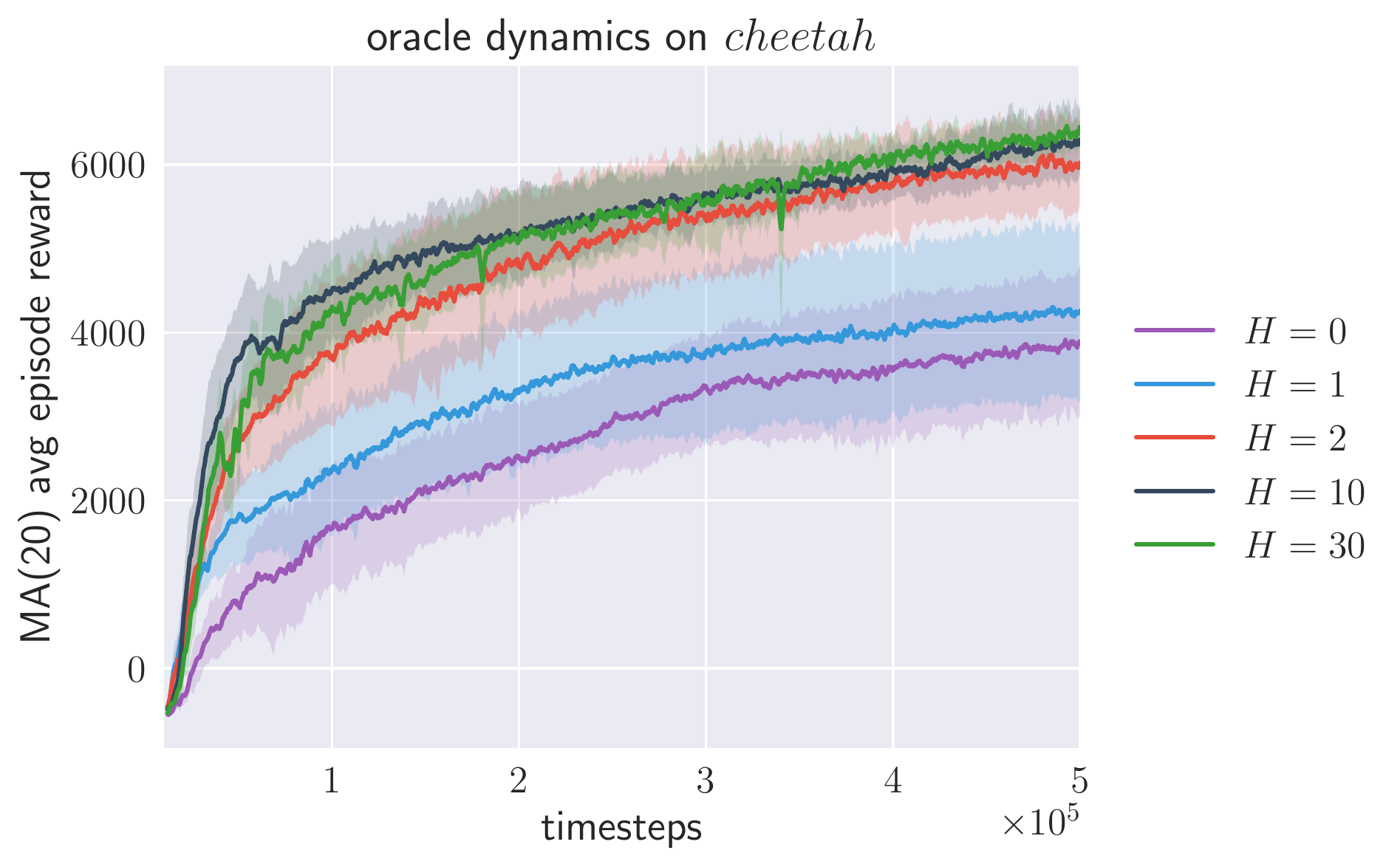}} \subfigure[]{\includegraphics[width=0.9\columnwidth]{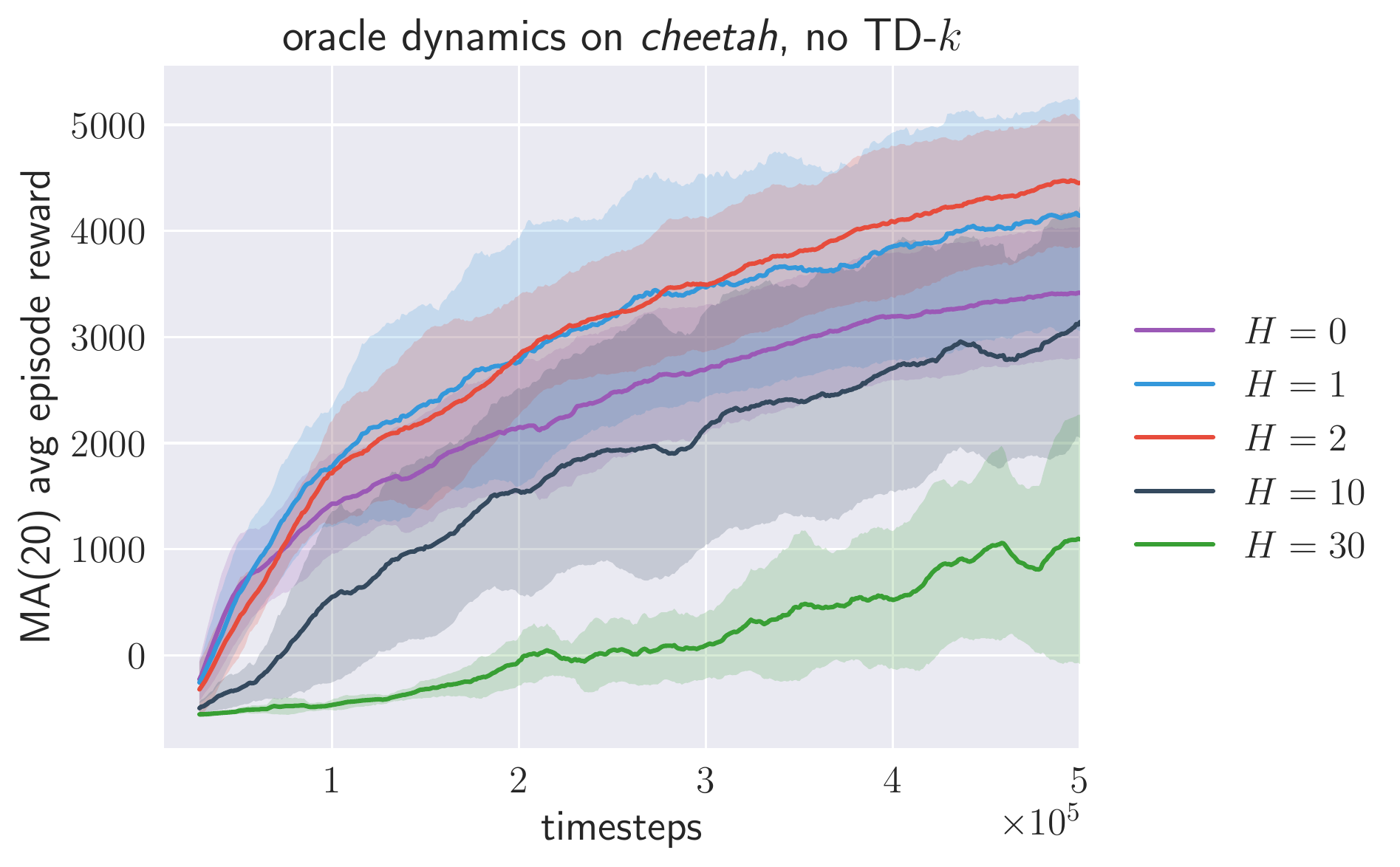}}
\end{center}
\caption{Learning curves for the \textit{cheetah} environment for MVE-DDPG with an ideal, oracle dynamical model at different horizons $H$ of model prediction. We examine performance (a) with and (b) without the TD-$k$ trick. $H=0$ implies no model use; this is the original DDPG. First, (a) exemplifies that improving value estimation with a model has a marked effect on performance in dense reward environments and offers an upper bound to the improvement that can result from learned-dynamics MVE. Note that as mentioned in Fig.~\ref{fig:qdensity} the batch size for oracle dynamics evaluations was reduced out of computational necessity, so these curves are not comparable to Fig.~\ref{fig:learned-hc-hs}. The diminishing returns for increases in $H$ that we observe further emphasize that model improvement is captured even with a short horizon. Second, (b) demonstrates the value of the TD-$k$ trick: for small $H$, distribution mismatch is small so (b) still shows a performance improvement, but as $H$ increases we lose the monotonic improvements observed in (a).
}
\label{fig:hc-true}
\end{figure*}

In addition, we verify that the TD-$k$ trick is essential to training $\hat Q$ appropriately. To do so, we conduct an ablation analysis on the \textit{cheetah} environment: we hold all parameters constant and substitute the learned dynamics model $\hat f$ with the true dynamics model $f$, making model error zero. If $\MSE_{\beta}(\hat Q)\approx \MSE_{(f^\pi)^H\beta}(\hat Q)$, the MVE estimates must improve exponentially in $H$, even without the TD-$k$ trick. However, this is not the case. With the TD-$k$ trick, increasing $H$ yields increasing but diminishing returns (Fig.~\ref{fig:hc-true}). Without the adjustment for distribution mismatch, past a certain point, increasing $H$ hurts performance. Because the dynamics model is ideal in these cases, the only difference is that the critic $\hat Q$ is trained on the distribution of states $\nu=\frac{1}{H}\sum_{t=0}^{H-1}(f^\pi)^t\beta$ instead of $\beta$, where $\beta$ is the empirical distribution resulting from the replay buffer. Since the TD-$k$ trick increases the support of the training data on which $\hat Q$ is trained, the function class for the critic may need to have sufficient capacity to capture the new distribution, but we did not find this to be an issue in our experiments.
 

\section{Related Work}

A number of prior works have sought to incorporate dynamics models into model-free RL and value estimation. We observe three general approaches: (1) direct incorporation of dynamics into the value gradient, (2) use of imagination as additional training data, and (3) use of imagination as context for value estimation.

(1) Stochastic value gradients (SVG) uses its model for improved credit assignment from traces of real trajectories \cite{svg}. By applying a differentiable dynamics model to real data only, SVG avoids instability from planning with an overfitted dynamics model, since the model is used for its gradients rather than its forward predictions. A major limitation of this approach is that the dynamics model can now only be used to retrieve information that is already present in the observed data, albeit with lower variance, so the actual improvement in efficiency is relatively small.
Applied to a horizon of length one, given off-policy data from $\beta$, $\mathrm{SVG}(1)$ estimates $J_\beta(\pi)\approx \E\ha{w(A|S)\pa{r(S,A) + \gamma \hat V(\hat S)}}$ with $S,A,r(S,A)$ observed from the transition taken according to some sampling distribution $\beta$ and $\hat S$ following the probabilistic dynamics model (the stochastic value gradient is achieved by using the reparameterization trick to differentiate the approximation). \citet{svg} does not consider the case of $\mathrm{SVG}(n)$ for off-policy data and $n>1$, likely due the use of the importance weight $w(a|s)=\frac{\pi(a|s)}{\beta(a|s)}$, which may vanish for model-based expansion of longer lengths. This is not a problem for the on-policy $\mathrm{SVG}(\infty)$, but by the authors' own account $\mathrm{SVG}(\infty)$ is less sample-efficient than $\mathrm{SVG}(1)$.

(2) Alternatively, Dyna-like approaches use the learned model for imagination rollouts \cite{dyna}. Providing imagined data to model-free value estimation algorithms is potentially a more robust use of a potentially erroneous view of a task's dynamics compared to planning. A number of follow-up methods have expanded on this idea.
For instance, in model-based acceleration (MBA), imagination rollouts are used as additional training data for a parameterized value network \cite{naf}. The original MBA proposal adds a replay buffer with the imagined data. Imagined states may then be used as the starting point for further imagination at later iterations. This reuse violates the rule of trusting the model for only $H$ steps of simulation. The authors find that this approach does not work well with neural network models, and settle for locally linear ones. Even with this modification, the model is deactivated in the middle of training heuristically to attain the best results. Our analysis and comparisons shed light on why this approach may be ineffective.

One approach to limit model use automatically is model-assisted bootstrapped DDPG (MA-BDDPG) \cite{kalweit}. This has two modifications from MBA. The first is that the imaginary states are not used as the starting point for future imaginary rollouts (this change alone is referred to as MA-DDPG). The second change adds an estimate of approximate critic uncertainty via the bootstrap. The imagined states are then used for training with frequency increasing in estimated critic uncertainty. Thus model use is limited except in scenarios where it is needed disambiguate uncertainty in the model-free critic estimate. Fig.~\ref{fig:learned-hc-hs} compares MVE to the IB approach of MA-DDPG. We do not use bootstrap to limit model use as MA-BDDPG does. The IB approach is in part representative of MBA as well, though MBA explicitly states that it is not compatible with neural network dynamics. MBA also uses the model for local planning in its imagination rollouts, and this is not done here.

Another prior approach is model-ensemble trust region policy optimization (ME-TRPO) \cite{metrpo}. Here, real data is used to train the dynamics model only, and a TRPO algorithm learns from imagination rollouts alone. ME-TRPO limits over-training to the imagined data by using an ensemble metric for policy performance. In other words, ME-TRPO bootstraps the dynamics model prediction, as opposed to the critic as in MA-BDDPG. However, dynamics model predictions are treated equally regardless of imagination depth, and the lack of use of real data for policy training may become problematic if dynamics models fail to capture phenomena occurring in reality. In any case, ME-TRPO demonstrates the success of using the bootstrap for estimating the degree of uncertainty in models, and we believe that this notion can be incorporated in future work combined with MVE. ME-TRPO is orthogonal and complementary to our method.

(3) The use of imagination as context for value estimation is most similar to MVE, but existing literature does not address model accuracy. In particular, unlike the aforementioned works, which focus on quantification of model uncertainty to limit model use, an end-to-end approach, I2A, avoids explicit reasoning about model inaccuracy \cite{i2a}. I2A supplies imagined rollouts as inputs to critic and actor networks, which are free to interpret the imaginary data in whatever way they learn to. A related approach is proposed by value prediction networks (VPN), which expand encoded state predictions and perform backups on a set of expanded imagined paths to compute an improved target estimate \cite{vpn}. Both I2A and VPN are tested on planning problems in discrete spaces. However, in continuous spaces, some degree of dynamics prediction error is unavoidable and may affect I2A stability or sample complexity as the network must learn to deal with uncertain rollouts. In addition, VPN crucially relies on the ability to perform stable backups on imagined paths with respect to current and future actions. These backups amount to optimizing over the action space to maximize the value function, which may be problematic in the continuous case. However, we believe that a stable model-based value expansion, which may be achieved with the TD-$k$ trick, along with careful optimization, may make a VPN-like approach viable in continuous contexts.

Finally, we note that MVE has some high-level similarity to $n$-step return methods. Both use short horizon rollouts to improve the value estimate, frequently used as a target value in computing Bellman error. Usually, $n$-step return methods are on-policy: they estimate target $Q$-values at some state-action pair as $\sum_{t=0}^{H-1}\gamma^tr_t+\gamma^H\hat Q(s_H,a_H)$ for a trace of rewards $r_t$ in an observed $H$-step trajectory \cite{inc}. The main difference of MVE from $n$-step return methods, explicit state prediction via dynamics modeling, is essential because it enables faster learning by use of off-policy data.
% https://link.springer.com/article/10.1007/BF00114731
Recent work, path consistency learning (PCL), relieves the on-policy requirements: PCL trains on a soft consistency error that is compatible with off-policy trajectories \cite{pcl}. % https://arxiv.org/abs/1702.08892
We believe that model-based extension of PCL with imagined rollouts may be possible following the algorithm design lessons MVE recommends. In particular, using an imagination buffer of rollouts to augment paths chosen for consistency learning may result in stale data and requires model use to be tempered by the stability of the algorithm to taking many gradient steps without gathering additional data. An MVE approach could augment paths with imagined branches that would be followed by the current policy, and the path consistency value would be the average over such branches. In other words, MVE may be complimentary to PCL, similar to how off-policy actor-critic methods can still be improved with MVE value estimation from imagined data, even though they are already off-policy.

\section{Conclusion}

%At face value, it is questionable whether robotics environments are an appropriate fit for the deterministic MDP model.
%%SL.02.08: can probably omit the philosophy here... not sure it really has anything to do with the purpose of experiments
%In reality, there are sources of stochasticity in continuous control stemming from imperfect actuation and observation measurement error. Such noise is usually are heteroschedastic and autocorrelated, which makes it difficult to simulate.
%We posit that a deterministic physics model suffices for the underlying physical phenomena and short horizons since we need to create a representative imagination of future states and rewards for learning rather than a robust one for planning. The assumption that the reward is explicitly known is reasonable here, as the reward function is typically manually designed.

In this paper, we introduce the model-based value expansion (MVE) method, an algorithm for incorporating predictive models of system dynamics into model-free value function estimation. Our approach provides for improved sample complexity on a range of continuous action benchmark tasks, and our analysis illuminates some of the design decisions that are involved in choosing how to combine model-based predictions with model-free value function learning.

Existing approaches following a general Dyna-like approach to using imagination rollouts for improvement of model-free value estimates either use stale data in an imagination buffer or use the model to imagine past horizons where the prediction is accurate. Multiple heuristics \cite{metrpo,kalweit} have been proposed to reduce model usage to combat such problems, but these techniques generally involve a complex combination of uncertainty estimation and additional hyperparameters and may not always appropriately restrict model usage to reasonable horizon lengths. MVE offers a single, simple, and adjustable notion of model trust ($H$), and fully utilizes the model to that extent. MVE also demonstrates that state dynamics prediction enables on-policy imagination via the TD-$k$ trick starting from off-policy data.
 
Our work justifies further exploration in model use for model-free sample complexity reduction. In particular, estimating uncertainty in the dynamics model explicitly would enable automatic selection of $H$. To deal with sparse reward signals, we also believe it is important to consider exploration with the model, not just refinement of value estimates. Finally, MVE admits extensions into domains with probabilistic dynamics models and stochastic policies via Monte Carlo integration over imagined rollouts.

% 8 pages exc refs + acks, 10 including.
% No acks in first version
% appendix in separate place for first version.

\section*{Acknowledgements}

The authors are very thankful for the valuable feedback from Roberto Calandra, Gregory Kahn, Anusha Nagabandi, and Richard Liaw. This research is supported in part by DHS Award HSHQDC-16-3-00083, NSF CISE Expeditions Award CCF-1139158, and gifts from Alibaba, Amazon Web Services, Ant Financial, CapitalOne, Ericsson, GE, Google, Huawei, Intel, IBM, Microsoft, Scotiabank, Splunk and VMware.

\FloatBarrier
\bibliography{main}
\bibliographystyle{icml2018}

\clearpage

\FloatBarrier

\appendix

\section{Experiment Details}\label{sec:exp-det}

To simulate MVE behavior with an ideal, oracle dynamics model, we needed to use the true environment, the MuJoCo simulation, for dynamics predictions. To make experiments tractable we implemented our own \texttt{cython}-based interface to the physics engine, based off of \texttt{mujoco-py} \cite{mujoco}. However, we were able to use the usual \texttt{gym} interface when applying learned dynamics.\footnote{We based environments off of \texttt{gym} commit 4c460ba6c8959dd8e0a03b13a1ca817da6d4074f.}

All runs wait until $10\cdot 10^3$ timesteps to be collected until DDPG training starts. We wait until $5\cdot 10^3$ timesteps have been collected until model training begins. We perform 4 gradient steps on both the dynamics and DDPG networks per collected timestep, though more may be possible. Batch sizes were always 512, except when training on the oracle dynamics model, where they were reduced to 32 out of computational necessity. In all experiments with learned dynamics, we use 4 seeds. To make up for the increased variance in the runs, we ran all experiments with the true dynamics model with 6 seeds.

\begin{table}[h!]
\begin{center}
\caption{Final tuned DDPG parameters for all the environments. \textsc{LR} refers to learning rate. The networks had 2 hidden layers of 64 units each. Adaptive parameter-space noise was fixed at 0.2. Decay refers to target decay}\label{tbl:ddpg-params}
\begin{small}
\begin{tabular}{lcccr}
\toprule
\textsc{Environment} & \textsc{Actor LR}  & \textsc{Critic LR} & \textsc{Decay} \\
\midrule
\textit{cheetah} & $10^{-3}$ & $10^{-3}$ & $10^{-2}$ \\
\textit{swimmer} & $10^{-4}$ & $10^{-3}$ & $10^{-2}$ \\
\textit{walker} &  $10^{-4}$ & $10^{-4}$ & $10^{-3}$ \\
\bottomrule
\end{tabular}
\end{small}
\end{center}
\end{table}

\begin{table*}[t]
\begin{center}
 \parbox{0.8\textwidth}{
\caption{Descriptions of the environments used for testing, all of which are based off of their corresponding \texttt{gym} environments with the usual unmodified rewards for forward motion.}}
\label{tbl:env-description}
\begin{small}
\begin{tabular}{lcccc}
\toprule
\textsc{Environment} & \texttt{gym} \textsc{Base Environment}  & \textsc{Modifications} & {\textsc{Input Dimension}} & {\textsc{Output Dimension}} \\
\midrule
\textit{cheetah} & \texttt{HalfCheetah-v1} & Add $x$ coordinate & $18$ & $6$ \\
\textit{swimmer} & \texttt{Swimmer-v1} & Add $x,y$ coordinates & $18$ & $6$ \\
\textit{walker} & \texttt{Walker2d-v1} & Add $x$ coordinate & $10$ & $2$ \\
\bottomrule
\end{tabular}
\end{small}
\end{center}
\end{table*}

\section{Off-Policy Deterministic Continuous Policy Improvement Theorem}\label{sec:dpg-continuous}

\textbf{Definition}. A direction $\vv\in\R^n$ ascends $\Phi:\R^n\rightarrow\R$ at $\vx$, or is an ascent direction for $\Phi$ at $\vx$, if there exists an $\epsilon> 0$ such that for all $\epsilon'\in(0,\epsilon)$, $\Phi(\vx + \epsilon'\vv)>\Phi(\vx)$.

\textbf{Remark}. If $\nabla \Phi(\vx)$ exists and is nonzero then the ascent directions are exactly those directions having positive inner product with the gradient.

Let $\pi_\theta:\mcS\rightarrow \mcA$ be a deterministic policy for an MDP admitting a transition measure $f$ (which may be a delta measure, corresponding to deterministic dynamics) from $\mcS\times \mcA$ to $\mcS$, which are all open subsets of a Euclidean space. For any distribution of states $\beta$, let the off-policy objective be $J_\beta(\theta)=\E_\beta\ha{V^{\pi_\theta}(S)}$.

\textbf{Theorem} \cite{degris}. Define:
$$
g_\theta (s)= D_{\pi_{(\cdot)}(s)}(\theta)^\top \nabla_a Q^{\pi_\theta}(s, a)\big|_{a=\pi_\theta(s)},
$$
where $D_{\pi_{(\cdot)}(s)}(\theta)$ is the Jacobian of $\pi$ with respect to $\theta$ for a fixed $s$. To improve legibility, we use shortened notation $$
\nabla_aQ^{\pi_\theta}(s,a)\big|_{a=\pi_\theta(s)}=\nabla_a Q,$$
keeping in mind that the latter is implicitly a function of $\theta,s$. We require that $g_\theta(s)$ exists and is nonzero for $\beta$-almost every $s$. Then $\E_\beta\ha{g_\theta (S)}$ ascends $J_\beta$ at $\theta$.

\begin{proof}
Fix $s$ where $g_\theta(s)\neq \vzero$. Notice $g_\theta(s)$ is the gradient at $\varphi=\theta$ for the function
$$
h(\varphi)=\inner{\pi_\varphi(s),\nabla_aQ}.
$$
Then for $\epsilon$ sufficiently small if $\theta'=\theta+\epsilon g_\theta(s)$ then $h(\theta')>h(\theta)$. Then by a Taylor expansion of $Q^\pi(s,a)$ for fixed $s$ around $a$ at $a=\pi_\theta(s)$, for $\epsilon$ sufficiently small:
\begin{align*}
Q^{\pi_\theta}(s,\pi_{\theta'}(s)) &= Q^{\pi_\theta}(s, \pi_\theta(s))+ \inner{\nabla_aQ, \pi_{\theta'}(s)-\pi_\theta(s)} + o(\epsilon)\\
&= Q^{\pi_\theta}(s, \pi_\theta(s))+h(\theta')-h(\theta) + o(\epsilon)
\end{align*}
Now by the Taylor expansion of $h$ around $\varphi=\theta$, swallowing the error term into the existing one:
\begin{align*}
Q^{\pi_\theta}(s,\pi_{\theta'}(s)) &= Q^{\pi_\theta}(s, \pi_\theta(s))+\inner{g_\theta(s),\theta'-\theta} + o(\epsilon)\\
 &= Q^{\pi_\theta}(s, \pi_\theta(s))+\epsilon \norm{g_\theta(s)}^2 + o(\epsilon)\\
 &> Q^{\pi_\theta}(s, \pi_\theta(s))
\end{align*}
The last inequality holds only for $\epsilon$ sufficiently small. Now we may proceed in a manner nearly identical to \citep{degris}.

\begin{align*}
    J_\beta(\theta)&=\int \d{\beta(s)} V^{\pi_\theta}(s)=\int \d{\beta(s)} Q^{\pi_\theta}(s, \pi_\theta(s))\\
    &<\int \d{\beta(s)} Q^{\pi_\theta}(s, \pi_{\theta'}(s))\\
    &=\int \d{\beta(s)} \pa{r(s,\pi_{\theta'}(s))+\gamma \int\d{f(s'|s,\pi_{\theta'}(s))}V^{\pi_\theta}(s')}
\end{align*}

Notice we may continue unrolling $V^{\pi_\theta}(s')<r(s',a')+\gamma \int\d{f(s''|s',a')}V^{\pi_\theta}(s'')$ where $a'=\pi_{\theta'}(s')$. Continuing in this manner ad infinitum, with $a_t=\pi_{\theta'}(s_t)$:
\begin{gather*}
    J_\beta(\theta)<\int\d{\beta(s_0)}r(s_0,a_0)+\gamma \int \d f(s_1|s_0, a_0)\\
\;\;\;\;\;\;\;\;\;\;\;\;\;\;\;\;\;\;\;\;\times\pa{r(s_1,a_1)+\int \d f(s_2|s_1,a_1)\cdots}
\end{gather*}
At this point, one may notice that the right-hand side is $J_\beta(\theta')$.
\end{proof}

\textbf{Remark}. The requirement that $g_\theta(s)$ is almost surely nonzero is not an artificial result of the analysis: the DPG ascent direction requires local improvement everywhere for sufficiently small step size. Consider the following one-dimensional hill-climbing MDP with $\mcS=[-1,2]$. An agent attempts to climb the reward function $r(s)=s^3\indicator\ca{s<0}+\Phi(s-1)$, where $\Phi$ is a quadratic bump function.
$$\Phi(s)=\begin{cases}\exp\pa{\frac{-x^4}{1-x^2}}&x\in(-1,1)\\0&\text{otherwise}\end{cases}$$
We consider the linear policy class $\pi_\theta(s)= \theta$, where the action represents movement towards a state $\theta$ the agent finds desirable: the dynamics are given by agent movement from position $s$ to position $s+\delta(\pi_\theta(s)-s)$.

\begin{tikzpicture}
  \begin{axis}[ 
  mark=none,
    xmin=-1.1,xmax=2.1,samples=400,
    ymin=-0.8,ymax=1.1,
    xlabel=$s$,
    ylabel={$r(s)$}
  ] 
    \addplot [
    mark=none,]{x^3 * (x <= 0) + ((x>0) && (x<= 2) ) * exp(-(x-1)^4 / (1-(x-1)^2))}; 
  \end{axis}
\end{tikzpicture}

Note that $r'(0),r''(0),r'''_-(0)=0,0,6$ ($r_-'''$ is the left derivative) while $r'(1+\epsilon),r''(1+\epsilon),r'''(1+\epsilon)=-\Theta(\epsilon^3), -\Theta(\epsilon^2),-\Theta(\epsilon)$. Now, consider a pathological $\beta$ with equal mass split between $0$ and $1+\epsilon$, for some small $\epsilon > 0$, and suppose the current $\theta =0$. Because the agent has no local derivative information at $s=0$, but a step to the left at $1+\epsilon$ improves its situation, the direction $\E_\beta g_0(s)$ actually worsens performance: indeed, $g_0(0)=0$ yet $g_0(1+\epsilon)=-\Omega(\epsilon^3)$, so any step in the direction $\E_\beta g_0(s)$ results in $\theta <0$. But because $\abs{r'''_-(0)}>\abs{r'''(1+\epsilon)}$ for $\epsilon$ sufficiently small, any step in this direction reduces overall performance $J_\beta$!

\section{Model-based Value Expansion Error Theorem}\label{sec:mve-error}

\textbf{Theorem}. Let $s_0\sim\nu$ and define $s_t,a_t,r_t$ to be the states, actions and rewards that come about from acting according to $\pi:\mcS\rightarrow\mcA$ in the true environment with dynamics $f$. Let $\hat s_t,\hat a_t,\hat r_t$ be constructed from starting at $s_0$ and using the learned dynamics $\hat f$ to imagine an $H$-step rollout. If the model has $\epsilon$ generalization error at depth $H$, i.e., $\max_{t\in[H]}{\E\norm{\hat s_t- s_t}^2}\le \epsilon^2$, and $r,V^\pi$ are $L_r,L_V$-Lipschitz functions of state, then
$$
\MSE_\nu(\hat V_H)\le c_1^2\epsilon^2+ (1+c_2\epsilon)\gamma^{2H}\MSE_{(\hat f^\pi)^H\nu} (\hat V)\,,
$$
where $c_1,c_2$ grow at most linearly in $L_r,L_V$ and are independent of $H$ for $\gamma <1$. We assume $\MSE_{(\hat f^\pi)^H\nu} (\hat V)\ge \epsilon^2$ for simplicity of presentation, but an analogous result holds when the critic outperforms the model.

Recall we have the induced policy dynamics $f^\pi:\mcS\rightarrow\mcS$ from $f:\mcS\times \mcA\rightarrow\mcS$ where $f^\pi(\cdot)=f(\cdot,\pi(\cdot))$ (with analogous notation for $\hat f^\pi$). We denote the Lipschitz constants of $r(\cdot, \pi(\cdot))$ and $V^\pi$ as $L_r,L_V$.

\begin{proof}
Fix $s_0$. Recall Eq.~\ref{eq:mve}:
$$
\hat V_H(s_0)=\sum_{t=0}^{H-1}\gamma^t \hat r_t+\gamma^H\hat V(\hat s_H)
$$
Then we may decompose $(\hat V_H(s_0)- V^\pi(s_0))^2$ as:
$$\pa{\pa{\hat M-M}-\gamma^H\pa{\hat V(\hat s_H)- V^\pi(s_H)}}^2,$$
where $M=\sum_{t=0}^{H-1}\gamma^t r_t$ denotes the model-based component, defined analogously for the imagined $\hat M$.
We will then further rely on the decomposition $\hat M - M=\sum_{t=0}^{H-1}\gamma^t\pa{\hat r_t-r_t}$.

We will require repeated appeals to the Cauchy-Schwarz inequality. For any $L^2$ random variables $A,B$, notice:
\begin{align*}
\E \ha{(A-B)^2}&=\E A^2-2\E\ha{AB}+\E B^2\\
&\le \E A^2 + 2\E^{1/2}A^2\E^{1/2}B^2+\E B^2
\end{align*}

Since each of the aforementioned decompositions results in a difference of squares, we can bound the MSE:
\begin{align*}
\MSE_\nu \hat V_H&\le \E(\hat M-M)^2\\
&\quad +2\gamma^H\sqrt{\E(\hat M-M)^2\E(\hat V(\hat s_H)- V^\pi(s_H))^2}\\
&\quad +\gamma^{2H}\E(\hat V(\hat s_H)- V^\pi(s_H))^2\\
\end{align*}
Then we bound all model-based terms, taking $i,j$ over $[H]-1$:
$$
\E(\hat M-M)^2\le  \sum_{i,j}\gamma^{2(i+j)}\sqrt{\E\pa{\hat r_i-r_i}^2\E\pa{\hat r_j-r_j}^2}
$$
Next, by smoothness, $\hat r_i-r_i\le L_r\norm{\hat s_i-s_i}$ almost surely for all $i$, so $\E\pa{\hat r_i-r_i}^2\le L_r^2\E\norm{\hat s_i-s_i}^2$, so $\E(\hat M-M)^2\le \epsilon^2c_1^2$, where the constant is bounded by $c_1^2\triangleq\sum_{ij}\gamma^{2(i+j)}\le L_r^2\min\pa{H^2,\pa{1-\gamma^2}^{-2}}$.

Then the cross term can be simplified by the assumption that the critic peforms worse than the model (without this assumption, an additional $O(\epsilon)$ cross term remains).
\begin{gather*}
2\gamma^H\sqrt{\E(\hat M-M)^2\E(\hat V(\hat s_H)- V^\pi(s_H))^2}\le \\
2\gamma^H c_1\epsilon\sqrt{\E(\hat V(\hat s_H)- V^\pi(s_H))^2}\le \\
2\gamma^H c_1\E(\hat V(\hat s_H)- V^\pi(s_H))^2
\end{gather*}

At this point, we have for some $c_1'$ linear in $L_r$,
$$
\MSE_\nu \hat V_H\le c_1^2\epsilon^2+\gamma^{2H}(1+c_1')\E(\hat V(\hat s_H)- V^\pi(s_H))^2\,.
$$
To bound the remaining term we use the same technique yet again, decomposing
$$
 \hat V(\hat s_H)- V^\pi(s_H)=(\hat V(\hat s_H)- V^\pi(\hat s_H))- (V^\pi(s_H)- V^\pi(\hat s_H))
 $$
with Cauchy-Schwarz:
\begin{flalign*}
&\E(\hat V(\hat s_H)- V^\pi(s_H))^2\\
&\quad \le \E(\hat V(\hat s_H)- V^\pi(\hat s_H))^2\\
&\quad \quad +2\sqrt{\E(\hat V(\hat s_H)- V^\pi(\hat s_H))^2\E( V^\pi(\hat s_H)- V^\pi(s_H))^2}\\
&\quad\quad  +\E( V^\pi(\hat s_H)- V^\pi(s_H))^2.
\end{flalign*}
Since $\E(\hat V(\hat s_H)- V^\pi(\hat s_H))^2=\MSE_{(\hat f^\pi)^H\nu}\hat V$ we finish by bounding $\E( V^\pi(\hat s_H)- V^\pi(s_H))^2\le L_V^2\epsilon^2$ by relying on Lipschitz smoothness of $V^\pi$.\end{proof}
\end{document}